\documentclass{article}

\PassOptionsToPackage{numbers, compress}{natbib}
\usepackage[preprint]{neurips_2025}

\usepackage[utf8]{inputenc} % allow utf-8 input
\usepackage[T1]{fontenc}    % use 8-bit T1 fonts
\usepackage{hyperref}       % hyperlinks
\usepackage{url}            % simple URL typesetting
\usepackage{booktabs}       % professional-quality tables
\usepackage{amsfonts}       % blackboard math symbols
\usepackage{nicefrac}       % compact symbols for 1/2, etc.
\usepackage{microtype}      % microtypography
\usepackage{xcolor}         % colors
\usepackage{graphicx,subfigure} %figures
\usepackage{algorithm}
\usepackage{algpseudocode}
\algnewcommand{\algorithmicand}{\textbf{ and }}
\algnewcommand{\algorithmicor}{\textbf{ or }}
\algnewcommand{\OR}{\algorithmicor}
\algnewcommand{\AND}{\algorithmicand}
\algnewcommand{\var}{\texttt}

\usepackage{amsmath}
\usepackage{amsfonts}
\usepackage{enumitem}
\usepackage{wrapfig}

\title{The Pump Scheduling Problem: \\ A Real-World Scenario for Reinforcement Learning}

\author{%
  Henrique Donâncio\\
  Normandie Universit\'e, INSA Rouen\\
  LITIS\\
  Rouen, France \\
  \texttt{henrique.donancio@insa-rouen.fr} \\
  \And
  Laurent Vercouter \\
  Normandie Universit\'e, INSA Rouen\\
  LITIS\\
  Rouen, France \\
  \texttt{laurent.vercouter@insa-rouen.fr} \\
  \And
  Harald Roclawski \\
  Technical University of Kaiserslautern \\
  SAM \\
  Kaiserlaustern, Germany \\
  \texttt{roclawsk@mv.uni-kl.de} \\
}

\begin{document}

\maketitle

\begin{abstract}
Deep Reinforcement Learning (DRL) has demonstrated impressive results in domains such as games and robotics, where task formulations are well-defined. However, few DRL benchmarks are grounded in complex, real-world environments, where safety constraints, partial observability, and the need for hand-engineered task representations pose significant challenges. To help bridge this gap, we introduce a testbed based on the pump scheduling problem in a real-world water distribution facility. The task involves controlling pumps to ensure a reliable water supply while minimizing energy consumption and respecting the constraints of the system. Our testbed includes a realistic simulator, three years of high-resolution (1-minute) operational data from human-led control, and a baseline RL task formulation. This testbed supports a wide range of research directions, including offline RL, safe exploration, inverse RL, and multi-objective optimization. The code and data are available at: \url{https://gitlab.com/hdonancio/pumpscheduling}.
\end{abstract}
\section{Introduction}

The pump scheduling problem involves determining when to operate pumps to efficiently meet fluctuating water demands while respecting operational constraints. Scheduling strategies vary with the characteristics of the system, such as the network topology or variations in electricity prices. For example, in regions with variable energy prices, pumps can run during off-peak hours, with storage tanks supplying water during peak periods~\citep{candelieri2018bayesian}. Although energy dominates pump operation costs~\citep{nault2015lifecycle}, constraints such as limiting pump switching frequency to protect infrastructure and ensuring water storage turnover for quality must also be addressed.

Several optimization techniques, such as Bayesian Optimization (BO)~\citep{candelieri2018bayesian}, Branch-and-Bound (BB) algorithms~\citep{costa2016branch, menke2016exploring}, Genetic Algorithms (GA)~\citep{abiodun2013, luna2019improving}, and Model Predictive Control (MPC)~\citep{castelletti2023model}, have been applied to this problem. However, these approaches have important limitations. The population-based optimization of GA and the combinatorial search of BB incur high computational costs, restricting their application to small-scale networks using precomputed daily schedules based on historical water consumption patterns, without real-time adaptability. BO enables fine-grained decisions, but lacks dynamic foresight to consider the impact of actions in future states. MPC, while predictive, relies on accurate system models, which are vulnerable to unmodeled dynamics, hampering its robustness in complex networks. Recent works (see $e.g.$~\citep{Hajgato2020, seo21pump}) explore Deep Reinforcement Learning (DRL), highlighting its potential as a data-driven solution with advantages in real-time adaptability and scalability for complex water distribution systems.

Reinforcement learning (RL) (see details in~\citep{sutton18}) is a decision-making framework in which an agent learns from interactions with the environment to maximize cumulative returns. Some works have extended RL applications to diverse domains, including autonomous driving~\citep{liu21rlfd}, robotics~\citep{gu17robotic, nair2018overcoming}, video games~\citep{lample2017playing, mnih13, wurman2022outracing, vinyals2019grandmaster}, dialogue systems~\citep{jaques2019way}, and more~\citep{levine2020offline}. Despite its success, RL research is based primarily on virtual environments such as games~\citep{machado18revisiting} and control simulators~\citep{controlsuite}. In these virtual scenarios, the reward function that shapes the agent's objective(s) and the observation space that defines the agent's possible knowledge about the environment's state are given. In contrast, real-world problems often pose significant challenges, such as defining these representations to support agent decisions~\citep{nair2018overcoming}. Despite the potential of RL, relatively few real-world based scenarios are accessible to the research community.

This work aims to help bridge the gap between real-world applications and RL by introducing a testbed based on the pump scheduling problem in a water facility. We provide a real-world system's simulator and sensor's logged data regarding the pump operation, collected over three years at one-minute intervals from a human-led operation. Alongside these, we detail the operational constraints defined by hydraulic experts and describe the current human-led scheduling strategy. We also established a baseline representation of the RL task, including observation features and a reward function, and highlighted research opportunities enabled by these tools. Our goal is for this testbed to serve as a real-world benchmark for diverse RL approaches, including offline RL, safe exploration, inverse RL, multi-objective RL, and state representation learning. The main contributions are summarized below:

\begin{itemize}[leftmargin=10pt]
    \item We release an RL testbed grounded in real-world conditions, comprising a water distribution system simulator, three years of human-led operation data sampled at one-minute intervals, and the implementation of offline RL algorithms.
    \item We describe the operational characteristics of the water distribution system, including key constraints, the current human scheduling strategy, and a baseline RL task formulation that allows agents to minimize energy consumption while respecting safety and operational requirements.
    \item We benchmark state-of-the-art offline RL algorithms using the proposed simulator and task formulation, demonstrating that policies trained solely from logged data can surpass human-led operation in energy efficiency by up to $5\%$.
\end{itemize} 

\section{Water Distribution System Overview and Operational Constraints}\label{sec:wds_overview}

\begin{figure}[!ht]
    \centering
    \includegraphics[width=1\textwidth]{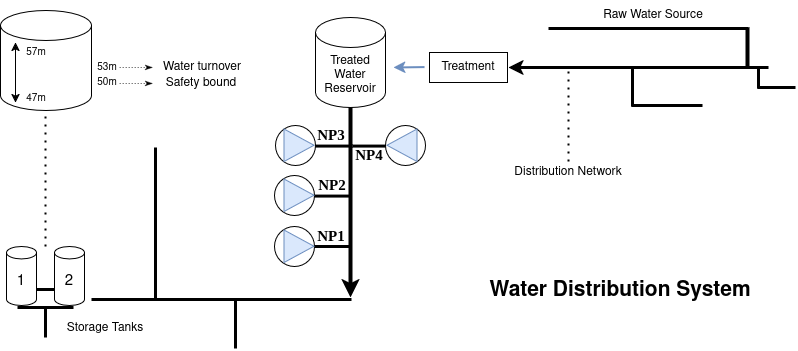} 
    \caption{Water distribution system overview. The system has four pumps with fixed speed (\texttt{ON}/\texttt{OFF}) and two elevated water storage tanks.}
    \label{fig:wds}
\end{figure}

Figure~\ref{fig:wds} provides an overview of the water distribution system analyzed in this work. The system draws and treats raw water from wells, which is then stored in a reservoir. In the water utility, four distribution pumps (NP1 to NP4) of varying capacities are available to pump water through the network to two elevated storage tanks. These pumps operate using start/stop control; speed or throttle control is not employed, and pumps operate individually, although parallel operation is used during periods of exceptionally high demand.

The pump scheduling considers multiple factors, including water demand forecasts, energy consumption, water quality, and supply security, to determine which pump is most suitable at any given time. The tanks are approximately $47$ meters above the pump station, have identical dimensions, and are effectively treated as a single tank with a total storage volume of $16\,000$ m³ and a maximum water level of $10$ meters. This results in a geodetic height differential of $47-57$ meters between the pumps and the storage tanks.

\begin{figure}
    \centering
    \subfigure[]{\includegraphics[width=.325\textwidth]{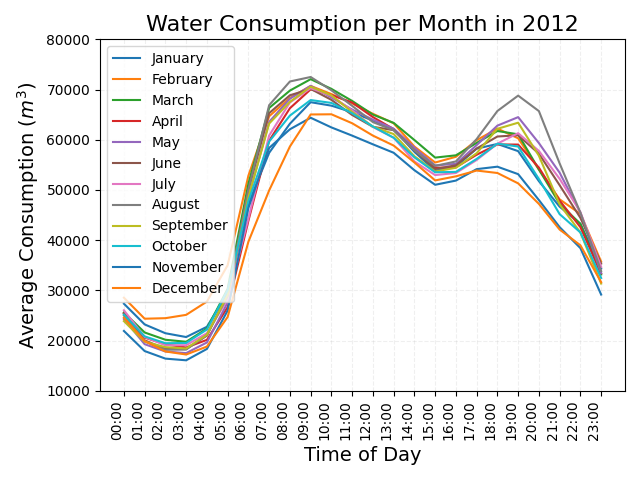}\label{fig:wca}} 
    \subfigure[]{\includegraphics[width=.325\textwidth]{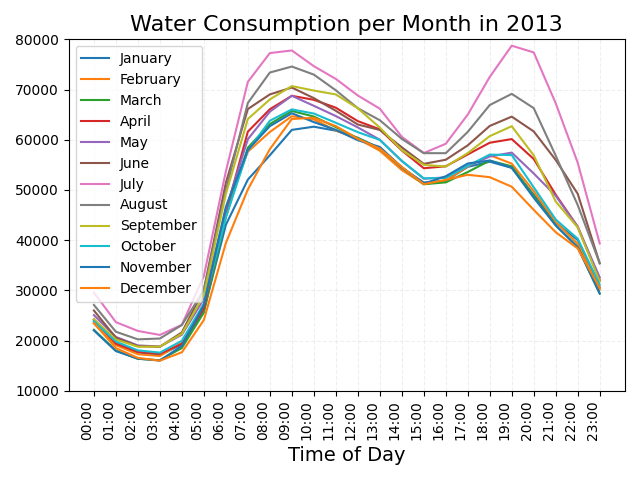}\label{fig:wcb}} 
    \subfigure[]{\includegraphics[width=.325\textwidth]{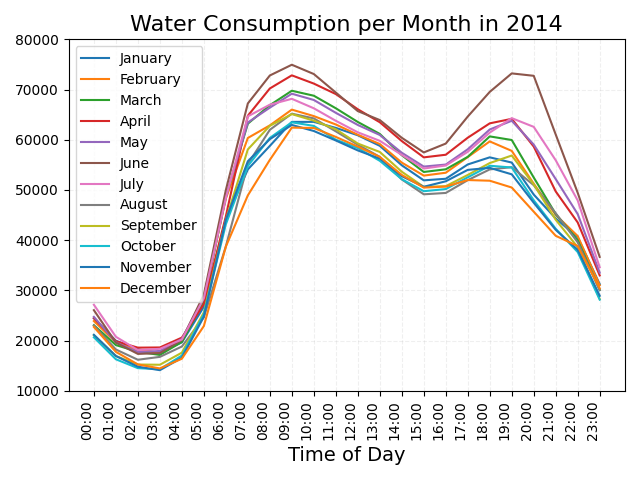}\label{fig:wcc}}
    \subfigure[]{\includegraphics[width=.325\textwidth]{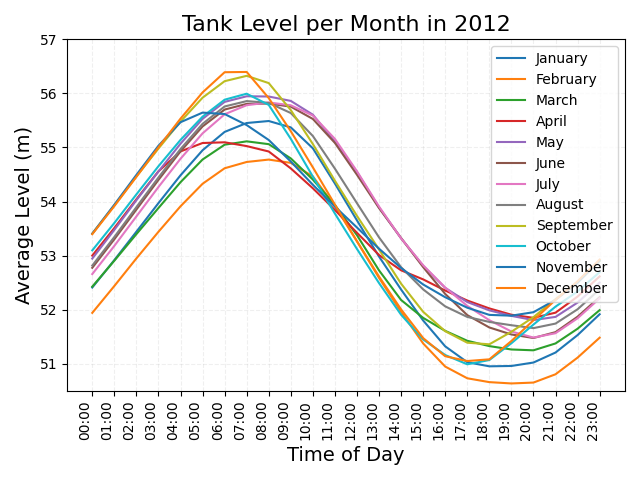}\label{fig:tka}}
    \subfigure[]{\includegraphics[width=.325\textwidth]{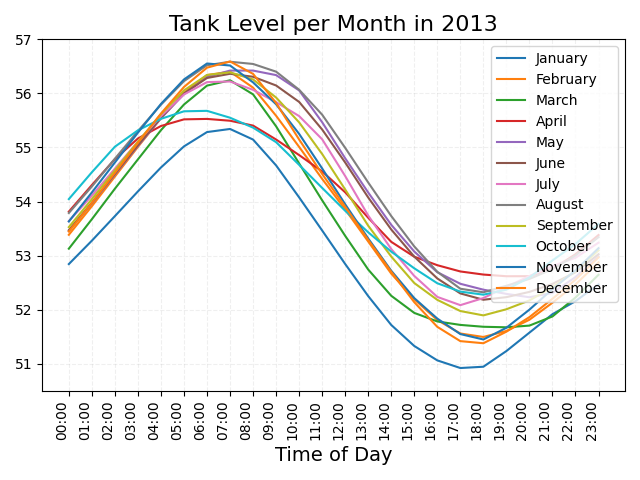}\label{fig:tkb}} 
    \subfigure[]{\includegraphics[width=.325\textwidth]{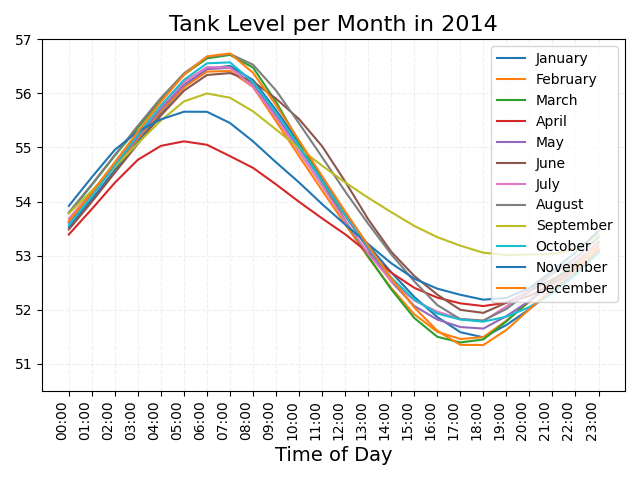}\label{fig:tkc}}
    \subfigure[]{\includegraphics[width=.325\textwidth]{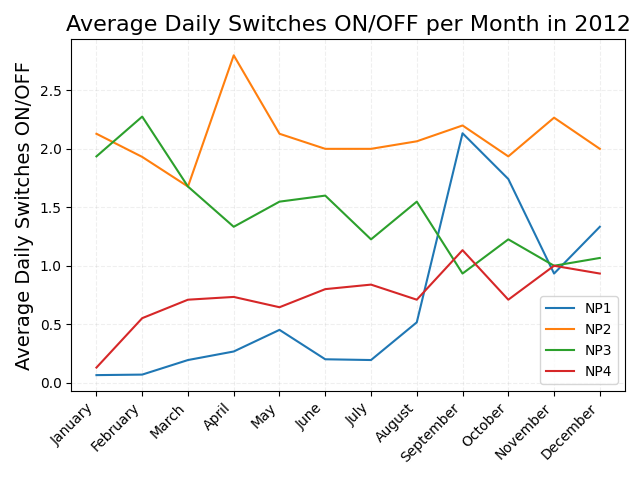}\label{fig:swa}}
    \subfigure[]{\includegraphics[width=.325\textwidth]{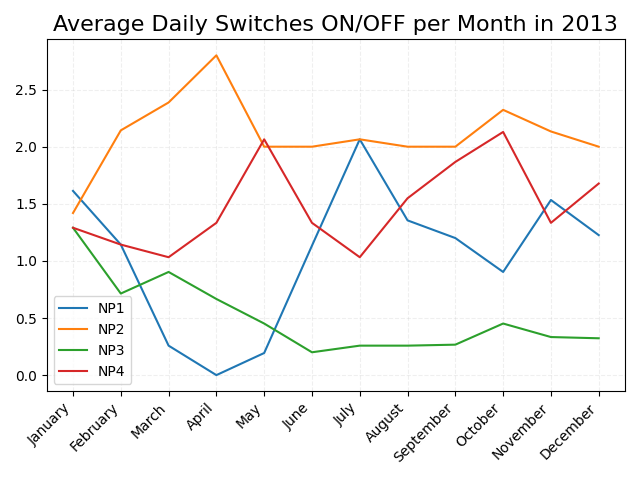}\label{fig:swb}}
    \subfigure[]{\includegraphics[width=.325\textwidth]{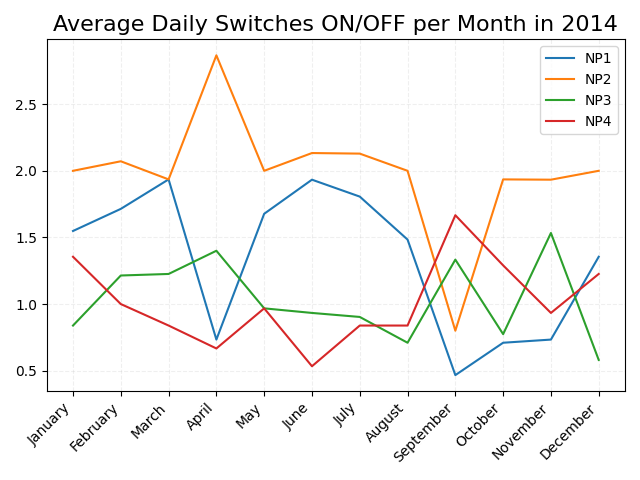}\label{fig:swc}}
    \subfigure[]{\includegraphics[width=.325\textwidth]{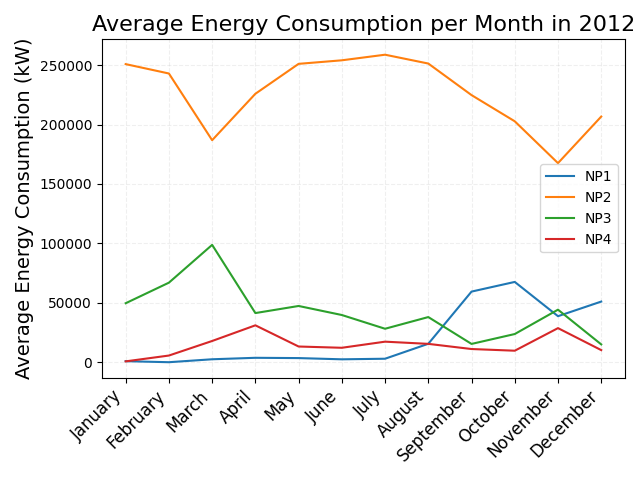}\label{fig:kwa}}
    \subfigure[]{\includegraphics[width=.325\textwidth]{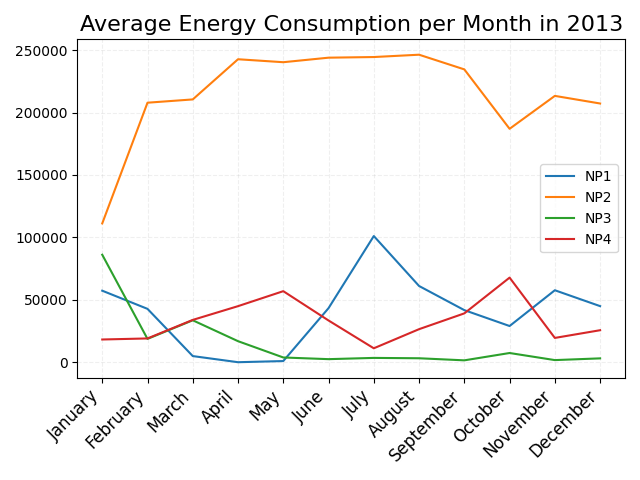}\label{fig:kwb}} 
    \subfigure[]{\includegraphics[width=.325\textwidth]{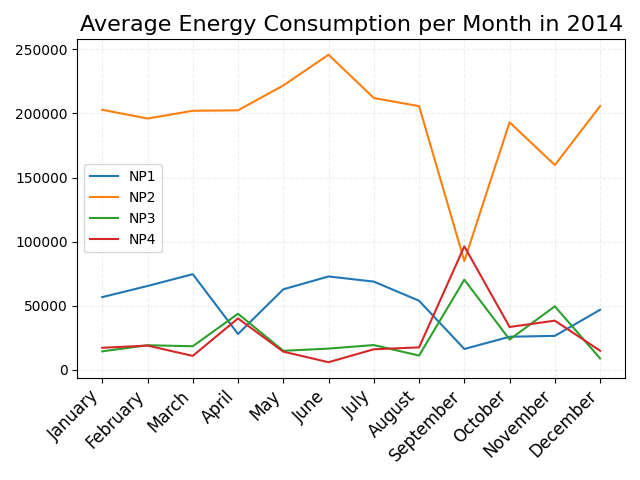}\label{fig:kwc}}
    \caption{Overview of human-led water system operation. 
(a–c) Water consumption. 
(d–f) Tank level profiles. 
(g–i) Pump switching. 
(j–l) Electricity usage.}
    \label{fig:behavior}
\end{figure}

In the water distribution system, human-led operation follows a strategy (policy) to handle pump scheduling. As shown in Figures~\ref{fig:tka},~\ref{fig:tkb},~\ref{fig:tkc}, the operation fills the tank to a high level before the peak in water consumption (see Figures~\ref{fig:wca},~\ref{fig:wcb},~\ref{fig:wcc}), and then they let it decrease throughout the day to provide water turnover in the tanks and keep its quality. Safety protocols require maintaining a minimum tank level of 3 meters to ensure operational robustness against unexpected events. Figures~\ref{fig:swa},~\ref{fig:swb},~\ref{fig:swc}, show the average daily pump switch per month. Each pump switch, either \texttt{ON} to \texttt{OFF} or \texttt{OFF} to \texttt{ON}, increments a counter by $+1$. Thus, we could say that the current pump operation generally uses each pump at most once a day. Although it is difficult to measure the impact of a strategy in preserving the system's assets, the idea is to minimize the amount of switching and provide a distribution of pump usage. Finally, in Figures~\ref{fig:kwa},~\ref{fig:kwb},~\ref{fig:kwc}, we show the electricity consumption where, given the pump settings, we see the prioritization of the pump NP2.

\section{The Pump Scheduling Problem as a Reinforcement Learning Task}\label{sec:pomdp}

Real-world RL problems, such as pump scheduling, often exhibit partial observability: the agent receives noisy or incomplete observations given the limited number of sensors and hidden variables influencing the system's dynamics. Theoretically, such problems are Partially Observable Markov Decision Processes (POMDPs)~\citep{sutton18}, with observations $o_t \in \mathcal{O}$ generated from states $s_t \in \mathcal{S}$ via an unknown observation function $\Omega: \mathcal{S} \mapsto \mathcal{O}$. However, practical approaches often cast these tasks as MDPs by approximating the Markov property. Techniques like frame stacking~\citep{mnih2015human} and recurrent neural networks (\emph{e.g.} LSTMs)~\citep{hausknecht15} enrich observations with temporal context, enabling agents to infer hidden dynamics from history. In this work, we adopt this approximation, defining the observation space $\mathcal{O}$ explicitly while noting that the observation function $\Omega$ remains unspecified. The task is thus formulated as a POMDP but approximated as an MDP with the tuple ($\mathcal{O}, \mathcal{A}, \mathcal{P}, \mathcal{R}, \gamma$), where:

\begin{itemize}
    \item $\mathcal{O}$ is the observation space that approximates the hidden state space $\mathcal{S}$;
    \item $\mathcal{A}$ is the action space;
    \item $\mathcal{P}: \mathcal{O} \times \mathcal{A} \times \mathcal{O} \mapsto [0, 1]$ is the transition probability, giving the likelihood of moving from observation $o$ to $o'$ after action $a \in \mathcal{A}$;
    \item $\mathcal{R}: \mathcal{O} \times \mathcal{A} \mapsto \mathbb{R}$ is the reward function;
    \item $\gamma \in [0, 1]$ is the discount factor, balancing immediate and future rewards.
\end{itemize}
    
The agent’s objective is to learn a policy $\pi:\mathcal{O}\mapsto\mathcal{A}$ that maximizes the expected discounted reward $J(\pi) = \mathbb{E}[\sum_{t=0}^{T-1} \gamma^t r(o_t, a_t)]$, where $T$ is the length of the episode and $r(o_t, a_t)$ is the reward given observation $o_t \in \mathcal{O}$ and action $a_t \in \mathcal{A}$. Based on the daily cycle of water consumption, we set $T = 1440$ timesteps, corresponding to one day of operation with decisions taken at 1-minute intervals. The observation space $\mathcal{O}$ includes sensor-derived features such as tank level and water consumption, while the action space includes four pumps, each with different flow rates $Q$ (m³/h) and power consumptions $kW$ (kW/h), and a no-operation option (NOP with $Q, kW = 0$). The reward function aims to supply water while maintaining safe tank levels and limiting energy use and pump switching. 

\paragraph{Observation space:} The observation at time $t$, denoted $o_t \in \mathcal{O}$, comprises the following features: the tank level $l_t \in [47, 57]$ (meters), the water consumption $c_t$ (m³/h) at time $t$, the time of day (step) $t \in [0, 1439]$ (minutes), the month $m \in {1, \ldots, 12}$, the previous action $a_{t-1} \in \mathcal{A}$, the cumulative run time of pumps $p_t \in [0, 1439]^{4}$ (minutes) and the turnover of the water tank $w(t) \in \{0, 1\}$ within the current episode. These features are designed to provide the agent with sufficient information about the hidden state to select actions that optimize the reward function. Observations are updated by simulating every 1-minute timestep, and the features are chosen to:

\begin{itemize}[leftmargin=10pt]
    \item Capture the current state of the system via the level of the tank ($l_t$) and the consumption of water ($c_t$);
    \item  Model temporal consumption patterns using the time of day ($t$) and month ($m$) to reflect daily and seasonal variations;
    \item Promote balanced pump usage and minimize switching by tracking the previous action ($a_{t-1}$) and cumulative runtime ($p_t$);
    \item Promote daily turnover in the water stored in the tanks to preserve its quality through the binary feature ($w_t$).
\end{itemize}
    
\paragraph{Action space:} The action space $\mathcal{A}$ represents the agent's control options on the pump system, defined as $\mathcal{A} = \{$ NP1, NP2, NP3, NP4, NOP$\}$. The actions correspond to activating one of four pumps, each with different flow rates ($Q$, in m³ / h) and power consumption ($kW$, in kW/h), ordered as NP1 > NP2 > NP3 > NP4 in both metrics, or selecting no operation (NOP), where $Q, kW = 0$.
 
\paragraph{Reward function:} The reward function, defined in Eq.\ref{eq:reward_function}, is designed to balance multiple subgoals: (i) maximize pump operation efficiency, (ii) maintain safe tank levels, (iii) penalize frequent pump switching and distribute usage across pumps, and (iv) promote daily water turnover in the tanks.

\begin{equation} \label{eq:reward_function}
r_t =
\begin{cases}
\exp\left(-\frac{kW_t}{Q_t}\right) - C \psi_t + Dw(t) - \ln\left(p_t(a_t) + P\right) & \text{if } Q_t > 0, \\ 
C\psi_t + Dw(t) - \ln\left(p_t(a_t) + P\right) & \text{if } Q_t = 0 \,,
\end{cases} 
\end{equation}

where:
\begin{itemize}[leftmargin=10pt]
    \item $\exp\left(-\frac{kW_t}{Q_t}\right)$ promotes energy efficiency by favoring higher flow $Q_t$ per power $kW_t$;
    \item $-C \psi_t$ enforces safety constraints, penalizing tank levels outside safe bounds;
    \item $D w_t$ encourages daily water turnover to maintain quality;
    \item $-\ln(p_t(a_t) + P)$ discourages prolonged pump use and frequent switching, where $p_t$ is the cumulative runtime of the active pump (selected by $a_t$) and $P$ is a switching penalty.
\end{itemize}

The constants $C$ and $D$ are weighting parameters (set both as $10$), and the terms $\psi_t$, $w_t$, and $P$ are defined as:

\begin{itemize}
    \item \textbf{Tank level constraint}:
    \begin{equation}
    \psi_t = \begin{cases} 
    \min\left(|l_t - 50|, 1\right) & \text{if } l_t < 50 \text{ (shortage risk)}, \\
    1 & \text{if } l_t = 57 \text{ (overflow)}, \\
    0 & \text{otherwise},
    \end{cases}
    \label{eq:psi}
    \end{equation}

    \item \textbf{Water turnover}:
    \begin{equation}
    w_t = \begin{cases} 
    1 & \text{if } w_{t-1} = 0 \text{ and } 50 \leq l_t < 53, \\
    0 & \text{otherwise},
    \end{cases}
    \label{eq:wt}
    \end{equation}

    \item \textbf{Switching penalty}:
    \begin{equation}
    P = \begin{cases} 
    1 & \text{if } a_t = a_{t-1} \text{ or } Q_t = 0 \text{ or } p_t(a_t) = 0, \\
    30 & \text{otherwise (pump switch)},
    \end{cases}
    \label{eq:P}
    \end{equation}
\end{itemize}

Although we show in the following sections that this task formulation can lead to policies that satisfy the objectives, its hand-crafted design may not ensure optimality~\citep{hayes21multi}, motivating future exploration of state representation and reward learning techniques discussed with further details in Section~\ref{sec:future}.

\section{Integrating the Water System Simulator into the RL Loop}\label{sec:simulator_loop}

\begin{figure}[!ht]
    \centering
    \includegraphics[width=.9\textwidth]{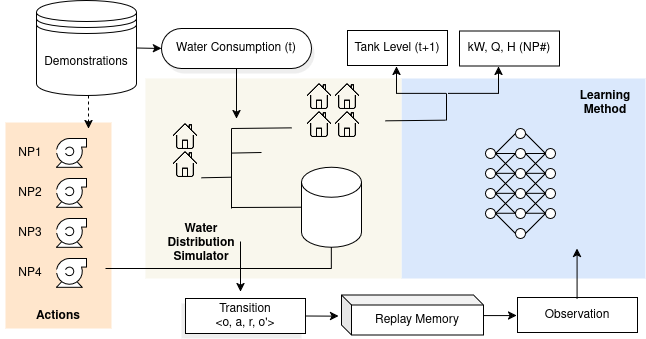} 
    \caption{The dynamics of a POMDP and the RL learning process through the water distribution system simulator.}
    \label{fig:rlsim}
\end{figure}

In RL, an agent selects an action $a$ given an observation $o$, receives a reward $r$, and transitions to a new observation $o'$. This process forms a transition tuple $\tau = \langle o, a, r, o' \rangle$, and a complete episode consists of a sequence of $T$ such transitions, \emph{i.e.} $\text{episode} = \sum_{i=0}^{T-1} \tau_i$. Figure~\ref{fig:rlsim} illustrates how this interaction loop is modeled for the pump scheduling problem using our simulator.

The simulator takes logged water consumption data at time $t$ as input, since this cannot be synthetically generated during new interactions. It also requires an initial condition for the tank level and other features that define the observation $o$. In the provided codebase, actions are derived from logged human data, although any control policy can be employed. A pump $\in \{\text{NP1}, \text{NP2}, \text{NP3}, \text{NP4}\}$ is considered \texttt{ON} if either its power consumption $kW_{NP\#}(t)$ or flow rate $Q_{NP\#}(t)$ are nonzero. If both values are zero for all pumps, the corresponding action is defined as NOP (no operation).

When an action $a$ is applied to the hidden state $s$, the simulator computes and returns the values of $Q_{NP\#}$, $kW_{NP\#}$, and $H_{NP\#}$ for the given timestep $t$ (see Appendix~\ref{app:simulator} for details). These outputs are then used to compute the reward $r(t)$. In the next timestep, a new tank level is observed, forming the next observation $o'$. This process creates a loop of transitions that are stored in a replay buffer~\citep{lin92experience}, following the Experience Replay paradigm. These collected transitions can be used to learn a new policy $\pi$, different from the human-led control strategy, by evaluating the demonstrations through the task representation proposed in Section~\ref{sec:pomdp}. To learn such a policy, we include some offline RL algorithms in our codebase, which are benchmarked in the following section.

\section{Benchmarking Offline RL on Real-World Demonstrations}\label{sec:benchmark}

In offline (batch) RL, the learning process is restricted to logged data, such as demonstrations gathered from human interactions with a system. This constraint often arises from the difficulty in creating accurate simulators or safety concerns related to real-world exploration~\citep{levine2020offline}. For example, an autonomous vehicle cannot employ trial-and-error learning in the environment due to safety risks and high costs. As a result, offline RL methods~\citep{fujimoto2019off, agarwal2020optimistic, kumar20conservativeq, jaques2019way} rely on the ability to exploit and generalize from static datasets to learn policies. Recently, benchmarks have been proposed to evaluate offline RL approaches in various scenarios~\citep{rlunplugged20, d4rl20, neorl22}. However, these scenarios often lack real-world constraints, such as safety, and the demonstrations, typically collected from multiple expert RL policies, offer sample diversity compared to datasets derived from a single source, such as human-led control.

\begin{wrapfigure}{r}{0.40\textwidth}
  \centering
  \includegraphics[width=0.38\textwidth]{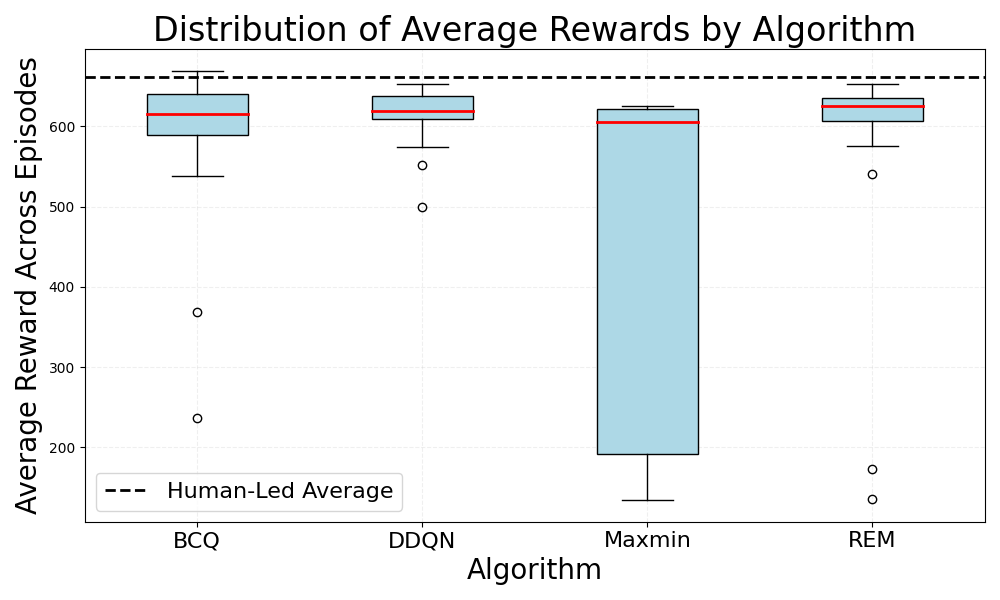}
  \caption{Benchmark results.}
  \label{fig:benchmark_results}
\end{wrapfigure}

In real-world tasks, the behavior policy, that is, the policy from which samples are derived, often exhibits a conservative approach considering exploration, and consequently, sample diversity. As illustrated in Figures~\ref{fig:tka},\ref{fig:tkb}, and~\ref{fig:tkc}, human control of pump operations follows a predictable pattern to manage tank levels. Consequently, critical scenarios such as overflow or water shortages are underrepresented in the logged data. This makes pump scheduling an ideal case for assessing the performance of offline RL algorithms using demonstrations from human interactions with real-world systems. To establish a baseline, we benchmark several algorithms (see Appendix~\ref{app:alg} for details)—BCQ~\citep{fujimoto2019benchmarking}, DDQN~\citep{van2016DDQN}, Maxmin Q-learning~\citep{lan20maxmin}, and REM~\citep{agarwal2020optimistic}—based on the experimental setup outlined in Appendix~\ref{app:setup}. Using three years of data, we present the results in Figure~\ref{fig:benchmark_results}, with one year of data on water consumption reserved for policy evaluation ($2012$) and two years of data allocated for training ($2013-2014$). The results reflect the average cumulative return across episodes, calculated using the reward function defined in Equation~\ref{eq:reward_function}.

To obtain the results shown in Figure~\ref{fig:benchmark_results}, we computed the average cumulative reward across 20 independently trained models for each offline RL algorithm. This was necessary to account for the inherent stochasticity of the learning process~\citep{henderson2018deep}. The dashed black line indicates the average cumulative reward of the behavior policy, \emph{i.e.}, the performance of the human-led control strategy evaluated under the proposed reward function. The results show that the median performance (red line) of all RL algorithms falls below the behavior policy. However, the best-performing models within each set remain competitive. This performance gap can likely be attributed to the challenges of offline RL: the learning process is restricted to demonstrations, making it difficult for policies to generalize beyond the observed state-action distribution. Algorithms such as BCQ and REM exhibit more consistent results, possibly because BCQ constrains action selection to logged data via behavior cloning, while REM stabilizes learning through full-network ensemble updates. Notably, DDQN achieves competitive performance, despite its lower computational burden, while the worst performance of Maxmin Q-learning might be explained by its overly conservative updates.

\begin{figure}[!ht]
\centering
\subfigure[]{\includegraphics[width=.245\textwidth]{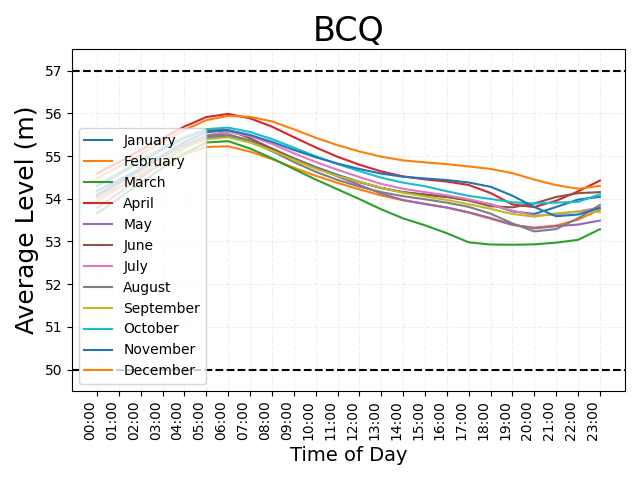}\label{fig:tank_bcq}}
\subfigure[]{\includegraphics[width=.245\textwidth]{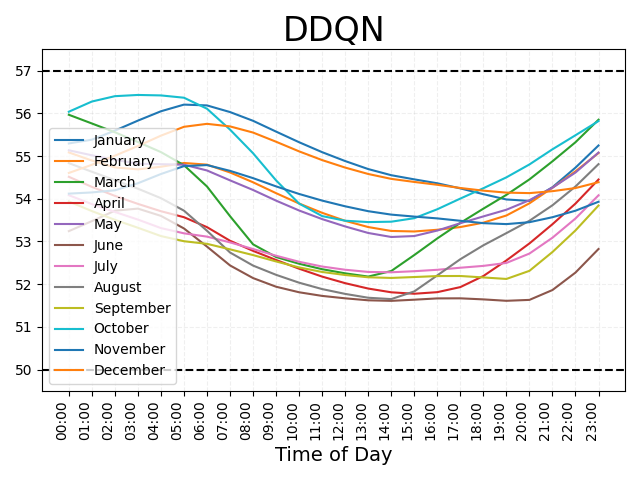}\label{fig:tank_ddqn}}
\subfigure[]{\includegraphics[width=.245\textwidth]{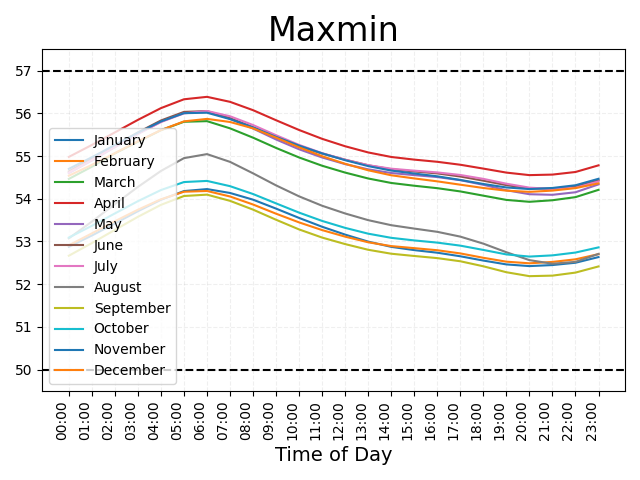}\label{fig:tank_maxmin}}
\subfigure[]{\includegraphics[width=.245\textwidth]{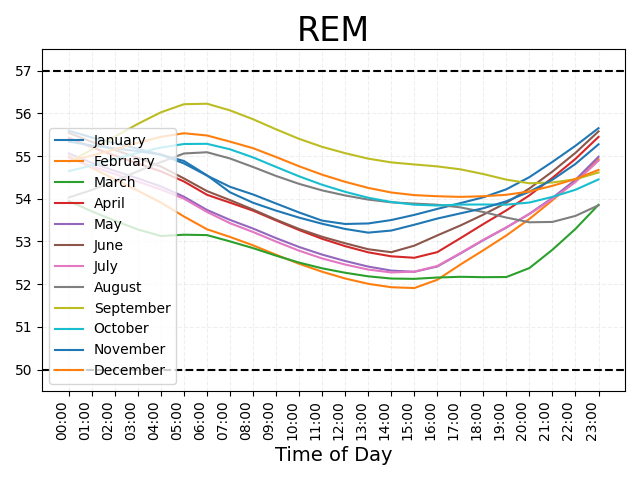}\label{fig:tank_rem}}
\subfigure[]{\includegraphics[width=.245\textwidth]{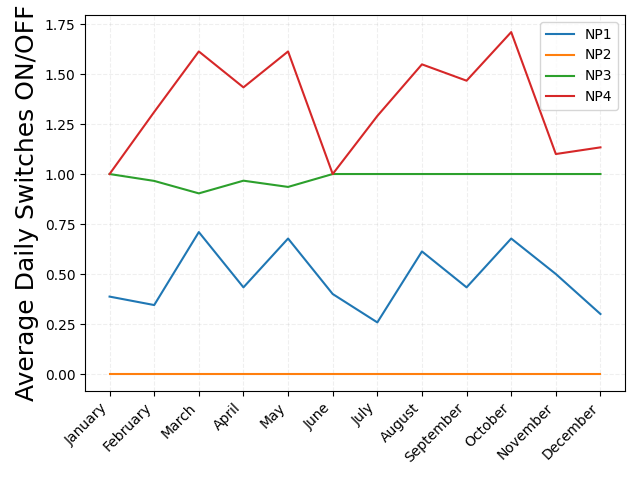}\label{fig:switch_bcq}}
\subfigure[]{\includegraphics[width=.245\textwidth]{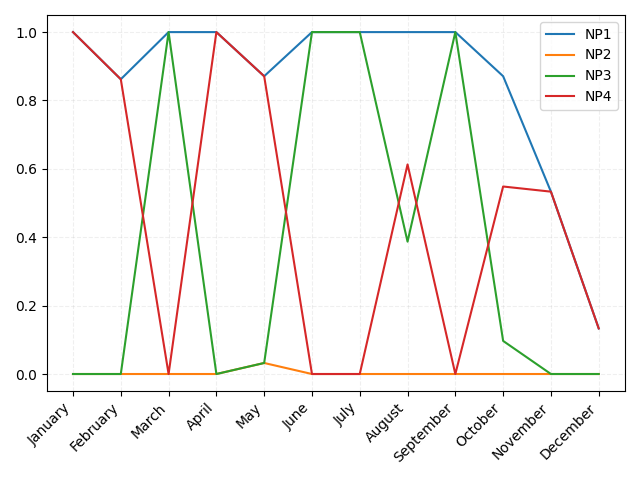}\label{fig:switch_ddqn}}
\subfigure[]{\includegraphics[width=.245\textwidth]{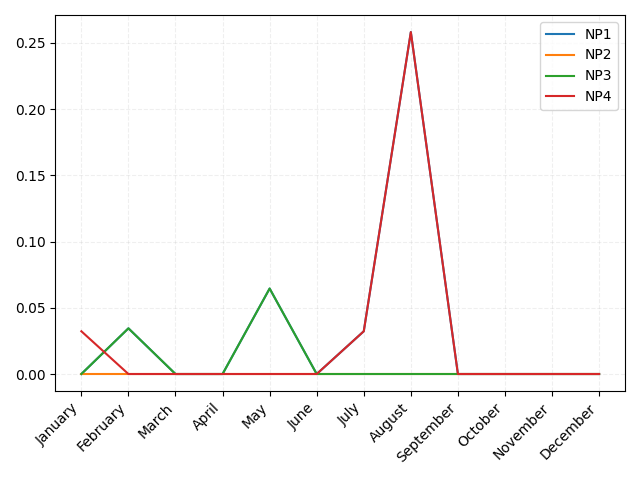}\label{fig:switch_maxmin}}
\subfigure[]{\includegraphics[width=.245\textwidth]{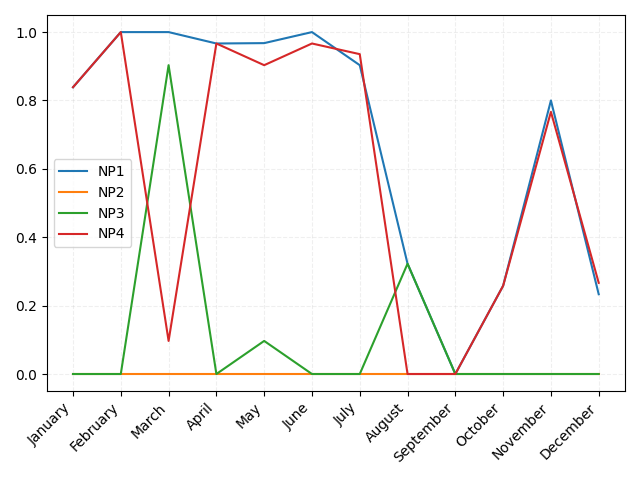}\label{fig:switch_rem}}
\caption{Tank levels (top row) and average daily pump switches (bottom row) for the best-performing policy of each offline RL algorithm: BCQ, DDQN, Maxmin Q-learning, and REM. All policies successfully maintain tank levels within safety limits while enforcing conservative switching patterns compared to human-led control.}
\label{fig:policy_behavior}
\end{figure}

Figure~\ref{fig:policy_behavior} illustrates the behavior of the best-performing policy (based on average cumulative reward) for each algorithm. As shown in Figures~\ref{fig:tank_bcq},\ref{fig:tank_ddqn},\ref{fig:tank_maxmin}, and~\ref{fig:tank_rem}, all learned policies maintained tank levels within safety bounds while minimizing pump switching frequency, often adopting more conservative behavior than the human-led baseline. In terms of energy consumption, the best policies achieved results comparable to or better than real-world control. Specifically, the relative energy savings compared to human-led control were: BCQ: $4.9\%$, DDQN: $-0.4\%$, Maxmin: $5.9\%$, and REM: $1.3\%$, where positive values indicate energy reduction relative to the expert baseline.

\section{Research Opportunities Enabled by the Testbed}\label{sec:future}

The pump scheduling testbed offers a rich environment for advancing several key research directions in RL. Below, we highlight opportunities across representation learning, safety-constrained exploration, inverse RL, multi-objective optimization, and control under continuous action spaces.

\paragraph{Representation learning and reward design.} Real-world problems like pump scheduling lack a predefined state representation or a reward function that produces optimal policies. In our testbed, the observation space was defined \emph{ad hoc}, based on the information (variables) provided by the sensors and the frequency at which these measurements are recorded. However, some variables may be uninformative or redundant, while incorporating additional variables or extending their historical context could improve policy efficiency. Some works (see \emph{e.g.}~\citep{lesort18state, merckling20srl, westphal2024variableselection}) have addressed the challenge of identifying more informative, low-dimensional state representations. Similarly, reward engineering poses difficulties, as defining a reward function that balances multiple, potentially conflicting objectives is complex. One potential solution is inverse RL~\citep{ng00inverse, abbeel04inverse}, which infers a reward function from the demonstrations. More recently, some approaches have completely bypassed the reward design by conditioning the learning process on achieving specific goals, such as reaching designated states (see \textit{e.g}.~\citep{ghosh2019learningTR}).

\paragraph{Multi-objective RL.} In real-world water distribution systems, the foremost priority is to guarantee a reliable supply of high-quality water to consumers. Once this requirement is satisfied, operational objectives such as minimizing energy consumption, reducing equipment wear, and managing risk, become relevant. These competing goals create a natural multi-objective optimization problem. However, in most RL applications, such objectives are aggregated into a single scalar reward using fixed weightings, masking the underlying trade-offs, and limiting flexibility under changing operational conditions. For example, consider a scenario in which planned maintenance is required. To proceed, operators must fill the storage tanks in advance to ensure uninterrupted supply. In such cases, a static reward function may fail to prioritize this temporary yet critical goal, as it cannot dynamically adjust to the objective importance change. Multi-objective RL (MORL)~\citep{roijers13morl, hayes21multi} addresses this limitation by representing rewards as a vector of cumulative returns, one for each objective. This enables explicit prioritization, allowing policies to adapt preferences over time without retraining.

\paragraph*{An exploration challenge.} One constraint in controlling the system is to avoid switching the pump operation too frequently to minimize mechanical wear and extend pump lifespan. However, exploration strategies such as \emph{epsilon-greedy}~\citep{sutton18} can lead to frequent action switches, which conflict with the need for repeated consecutive actions, potentially causing premature convergence to suboptimal policies. Moreover, approaches that prioritize visiting novel states~\citep{bellemare2016unifying, subramanian16exploration, tang2017exploration} risk violating safety constraints, such as tank level limits, by driving the system into unsafe configurations. Thus, the pump scheduling problem can be a setting for evaluating exploration strategies that aim to improve sample efficiency while limiting state visitation to those that meet system safety constraints~\citep{garcia15saferl}. In particular, our episodic formulation is \emph{state-persistent}, as the initial tank level of each episode depends on the final state of the previous episode, reflecting the continuity of the real world. This persistence of the state requires a robust policy to varying initial conditions, a challenge absent in most existing RL benchmarks~\citep{reyes20ecological}.

\paragraph*{Learning from demonstrations.} In this work, we evaluated the performance of offline RL algorithms, such as BCQ and REM, using a dataset of state-action pairs derived from expert demonstrations. These demonstrations reflect a consistent and safe human behavior, and consequently, a concentrated distribution: undesirable or risky system configurations, such as tank overflows or shortages, are rarely encountered. This makes the testbed particularly well-suited to studying offline RL challenges where generalization from narrow expert behavior is required. The testbed also enables the evaluation of Behavior Cloning (BC)~\citep{hussein17imitationsurvey} approaches, which correspond to solving a maximum likelihood problem to mimic the agent’s actions. BC can serve as a pre-training step to accelerate policy learning, although mitigating the distributional shift when the learned policies begin to generate new trajectories remains a critical issue~\citep{nair2018overcoming, nakamoto2023calql}.
Beyond imitation learning, demonstrations can also support alternative approaches, such as building dynamics models or guiding uncertainty-aware planning, as explored in model-based methods (see \emph{e.g.}~\citep{yu20mopo, kidambi2020morel}).

\paragraph*{Continuous action-space.} In its current configuration, the control space of the water distribution simulator is discrete: a finite set of pumps operate at fixed speeds, with the additional option of turning off all pumps. However, as discussed in Appendix~\ref{sec:continuous}, the simulator can be extended to support variable-speed pumps, enabling continuous control of flow rates. This opens the door to evaluating RL approaches designed for continuous action spaces, such as actor-critic methods~\citep{sutton18}. Such extensions would allow researchers to investigate smooth control policies with fine-grained actuation.

Together, these research directions underscore the testbed's versatility as a platform to advance RL research in realistic environments.

\section{Conclusion}

We introduce a real-world based RL testbed for the pump scheduling problem, containing a validated simulator and three years of high-resolution sensor data collected from human-led control. The simulator models system dynamics based on hydraulic principles, enabling end-to-end policy evaluation. We formally cast the problem as a POMDP, defining the observation and action spaces, reward function, and discussing the constraints to control the system, which guide the design of the task representation. Using the proposed POMDP, we benchmark state-of-the-art offline RL algorithms, showing that policies trained solely from demonstrations can match or exceed expert performance in energy efficiency while respecting safety requirements.

\paragraph{Limitations.}  
A key limitation lies in the finite nature of the dataset: unlike synthetic benchmarks such as Gymnasium~\citep{towers2024gymnasiumstandardinterfacereinforcement}, our testbed does not allow unlimited data generation, which restricts the training. However, this constraint reflects a common real-world challenge and positions the testbed as a platform for studying generalization, offline RL, and data-efficient learning. We view this work as a foundation for further development, and we encourage the community to expand the benchmark with additional challenges and settings not addressed in the current release.

% \begin{ack}
% The authors would like to acknowledge that Fig.~\ref{fig:wds} is based on an image created by Thomas Krätzig. The authors also acknowledge FAPEMIG, Federal Ministry of Education and Research of Germany, Agence Nationale de la Recherce de France, and Fonds de la Recherce Scientifique Belge, for funding this research by the project IoT.H2O (ANR-18-IC4W-0003) on the IC4Water JPI call.
% \end{ack}

%%%%%%%%%%%%%%%%%%%%%%%%%%%%%%%%%%%%%%%%%%%%%%%%%%%%%%%%%%

\bibliographystyle{unsrt}  
\bibliography{bibliografia} 

\pagebreak

\appendix

\section{Related Works}\label{sec:related}
Below, we outline related approaches in the literature on optimizing pump scheduling and control using RL.

\paragraph{Control using Reinforcement Learning.} In~\citep{ijcai2018-79} the authors use Deep Recurrent Q-Networks~\citep{hausknecht15} in the scenario of a smart grid. The objective is to develop a pricing strategy to maximize the broker's (agent) profits. A reward-shaping strategy is proposed once customers are clustered according to their consumption patterns and managed by their respective sub-brokers. Then, a mechanism for credit assignment was necessary to indicate the contribution of each sub-broker to the global return. Wei et al.~\citep{wei2017hvac} propose a \textit{data-driven} solution to control the HVAC (heating, ventilation, and air conditioning) system. The objective is to control the environment temperature by handling numerous disturbances with real-time data. The possible actions are a discrete set of airflow rates divided by building zone. Then, the approach splits each zone, controlling it with distinct neural networks. Sivakumar et al.~\citep{sivakumar19mvfst} propose a network control strategy that acts asynchronously with the environment. The delay, which the authors call $\delta$, corresponds to the policy lookup time when selecting an action. Meanwhile, the network transmits data in the interval [$t$, $t + \delta$]. Thus, the transitions depend on the state and action in some time step $t$ and the previous action in $t-1$. In~\citep{hay19congestion}, the problem of network congestion control is also addressed, and a testbed is released. In~\citep{bellemare2020autonomous}, RL is applied to a stratospheric balloon flight controller that must handle a partially observable environment and continuous interaction. Finally, Degrave et al.~\citep{degrave2022magnetic} use RL for nuclear fusion control, which achieved a variety of plasma configurations.

\paragraph{Pump Scheduling.} In~\citep{costa2016branch}, the authors adopted a branch-and-bound algorithm that interacts with the hydraulic simulator EPANET~\footnote{https://www.epa.gov/water-research/epanet} to evaluate decisions at each timestep corresponding to $1$ hour in a horizon length of $24$ hours. The proposed algorithm aims to find a solution with minimum electricity consumption. Menke et al. in~\citep{menke2016exploring} also used a branch-and-bound approach through an algorithm with several steps for the branching procedure. The objective function adopted is in contact with the one presented in this work. The authors consider pumps with fixed speed (\texttt{ON}/\texttt{OFF}), intending to minimize the power consumption while penalizing switches in pump operation. In~\citep{seo21pump}, the authors also applied DRL to control a real-world scenario of a wastewater treatment plant. Unlike our case study, in their system, the electricity price has different tariffs throughout the day, which adds another constraint to the proposed strategy. 

\section{Details of the Water Distribution System Simulator}

\subsection{Modeling Electricity Consumption with the Simulator}

\begin{figure}[!ht]
\centering
\subfigure[]{\includegraphics[width=.495\textwidth]{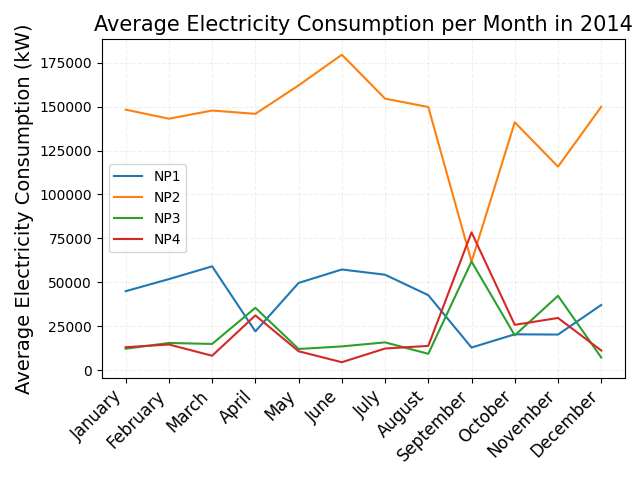}\label{fig:measured_kw}}
\subfigure[]{\includegraphics[width=.495\textwidth]{Figures/ElectricityConsumption2014.png}\label{fig:simulated_kw}}
\caption{Average electricity consumption calculated using measured data~\ref{fig:measured_kw} and simulator~\ref{fig:simulated_kw}.}
\label{fig:kw_comparison}
\end{figure}

Figure~\ref{fig:kw_comparison} shows the electricity consumption of the measured (real-world) data and the simulator. Note that the range of values between the measured and simulated data differs. The reason is that the pump efficiency is not considered in Equation~\ref{eq:pump_power}. Thus, we obtain the values for the measured data by taking the hydraulic power ($P_h$)  and dividing it by the efficiency $\eta$:

\begin{equation}\label{eq:pump_power}
    P_{h}(kW) = Q \rho g h / (3.6~10^{6}), \textrm{where}
\end{equation}

\begin{itemize}
    \item $Q$ is the flow $(m^{3}/h)$
    \item $\rho$ is the density of fluid $(kg/m^{3})$
    \item $g$ is the acceleration of gravity $(9.81 m/s^{2})$
    \item $h$ is the differential head (m)
\end{itemize}

 \subsection{Further Details on the Water System Simulator}\label{app:simulator}
 
The simulator illustrated in Figure~\ref{fig:wds} is developed to model the operational behavior of distribution pumps in a real-world water network facility located in Germany. The calculation of a pump's operating point is based on the intersection between its characteristic curve (hydraulic head versus flow rate) and the system curve.

The system curve represents the amount of hydraulic head required to deliver a given flow rate through the network. It consists of two components (Figure~\ref{fig:SystemCurve}): a static component, independent of flow rate, determined by the difference in geodetic height (\emph{e.g.} when pumping water into an elevated storage tank) and a dynamic component, which reflects friction losses in the piping system and increases quadratically with flow rate. The operating point of a pump is determined by the intersection of its hydraulic head curve with the system curve.

The system curve in this water distribution network is not fixed. The static head varies with the current tank level, while the slope of the dynamic component depends on hydraulic losses influenced by the instantaneous water demand. Higher water demand lowers the overall system pressure, resulting in a flatter system curve; conversely, lower demand increases system pressure, steepening the curve. The relationship between tank level, water demand, and system curve characteristics was derived from empirical measurements.

The simulator can model pump operations over arbitrary periods. The simulation process requires specifying the expected water demand for each time step, along with the initial tank level and the pump schedule. At each time step, the simulator computes the current system curve based on tank level and water demand, determines whether a pump is active, and calculates its operating point as the intersection between the pump and system curves. If the pump flow rate exceeds the water demand, the tank level increases accordingly; if the flow rate is insufficient, the tank level decreases. This process is repeated iteratively, updating the system curve and tank state at each step, to simulate continuous operation over the defined period.

\subsection{Logged Data: List of Variables}

Table~\ref{tab:variables} shows a list of variables included in the logged data to reproduce this work. The dataset contains three years in 1-minute timesteps of raw data gathered through sensors from a real-world system. This dataset provides, at some timestep $t$, information about the water tank level, and also pump information regarding electricity consumption \textit{kW}, water flow $Q$, and hydraulic head $H$.

\begin{table}[ht]
\centering
\caption{List of Variables in the Water Facility Dataset}
\label{tab:variables}
\small
\setlength{\tabcolsep}{6pt} % Adjust column spacing
\renewcommand{\arraystretch}{1.2} % Add vertical padding
\begin{tabular}{|l|l|l|}
\hline
\textbf{Variable Name} & \textbf{Units/Scale} & \textbf{Description} \\ 
\hline
\texttt{../Tank1/\{Month\}/`HB\_Niveau\_N\_pval'} & Meters & Water tank level \\ 
\hline
\texttt{../WaterConsumption/\{Month\}/`Netzverbrauch\_pval'} & m$^3$/h & Water consumption \\ 
\hline
\texttt{../NP/NP\{\#\}/KW/`NP\_\{\#\}\_SIMEAS\_P\_Leistung\_pval'} & kW/h & kW consumption \\ 
\hline
\texttt{../NP/NP\{\#\}/Q/`NP\_\{\#\}\_Volumenfluss\_pval'} & m$^3$/h & Water flow ($Q$) \\ 
\hline
\texttt{../NP/NP\{\#\}/H/`NP\_\{\#\}\_Druck\_Druckseite\_pval'} & Meters & Hydraulic head ($H$) \\ 
\hline
\end{tabular}
\end{table}

\section{Offline RL Algorithms Used in Benchmarking}\label{app:alg}

Learning policies constrained to expert demonstrations can be challenging due to the distributional shift issue and overly optimism in the face of uncertainty~\citep{fujimoto2019off, levine2020offline, ostrovski2021difficulty}. In this section, we further detail how the offline RL algorithms we benchmark tackle this issue.

\begin{itemize}[leftmargin=10pt]
    \item \textbf{Double DQN (DDQN)}~\citep{van2016DDQN} mitigates the overestimation bias present in standard Q-learning by decoupling action selection and evaluation during target computation. Specifically, the action that maximizes the Q-estimation is selected using the current network parameters $\theta_i$, but its value is estimated using the target network parameters $\theta_{i-1}$. DDQN is widely used as a baseline in both online and offline RL benchmarks. The loss $\delta_i$ used to train the Q-network is given by:

    \begin{equation}\label{eq:DDQN}
    \delta_{i} = \mathbb{E}\left[ R(s,a) + \gamma Q\left(s', \arg\max_{a'} Q(s', a'; \theta_i); \theta_{i-1} \right) - Q(s,a; \theta_i) \right]^2.
    \end{equation}

  \item \textbf{Random Ensemble Mixture (REM)}~\citep{agarwal2020optimistic} mitigates overestimation by averaging Q-values across an ensemble of estimators using randomly sampled convex combination weights. At each training step, a Dirichlet-distributed weight vector $\boldsymbol{\alpha}$ is sampled such that $\sum_{k=1}^{K} \alpha_k = 1$ and $\alpha_k > 0~\forall k$. This mixture is used to compute both the target and predicted Q-values. The loss $\delta_i$ is given by:

    \begin{equation}\label{eq:rem}
    \delta_{i} = \mathbb{E}\left[ R(s,a) + \gamma \max_{a'} \sum_{k=1}^{K} \alpha_k Q_k(s', a'; \theta^{k}_{i-1}) - \sum_{k=1}^{K} \alpha_k Q_k(s, a; \theta^{k}_i) \right]^2.
    \end{equation}

    \item Although not originally designed for offline settings, \textbf{Maxmin Q-learning}~\citep{lan20maxmin} mitigates overestimation bias by taking the minimum predicted Q-estimation across a set of Q-networks for each action: $Q^{\min}(s, a) = \min_{j \in \{1,\dots,N\}} Q_j(s, a)$, $\forall a \in \mathcal{A}$. During training, a random network $k \in N$ is randomly selected for each update step, but the target is computed using the minimum over all $N$ models:

    \begin{equation}\label{eq:maxmin}
    \delta_i = \mathbb{E}\left[ R(s,a) + \gamma \max_{a'} Q^{\min}(s', a'; \theta_{i-1}) - Q_k(s,a; \theta_i) \right]^2, \quad k \in \{1,\dots,N\}.
    \end{equation}

    \item \textbf{Batch-Constrained Deep Q-Learning (BCQ)}~\citep{fujimoto2019off, fujimoto2019benchmarking} mitigates overestimation by constraining the policy to select only actions that are likely under the behavior policy. In the discrete-action version, BCQ uses a BC model $G_\omega(a \mid s)$ trained on the dataset to estimate the action distribution. At each timestep, actions are filtered by a threshold $\tau$, and only those with a sufficiently high likelihood are considered for the target computation. The loss is defined as:

    \begin{equation}\label{eq:bcq}
    \delta_i = \mathbb{E} \left[ R(s,a) + \gamma \max_{a' \text{ s.t. } \frac{G_\omega(a' \mid s')}{\max_{\hat{a}} G_\omega(\hat{a} \mid s')} > \tau} Q(s', a'; \theta_{i-1}) - Q(s, a; \theta_i) \right]^2.
    \end{equation}

\end{itemize}

\section{Experimental Setup}\label{app:setup}

Reproducibility remains a major challenge in DRL research~\citep{henderson2018deep, neorl22}, often due to the lack of open-source code, limited access to training data, and insufficient documentation of implementation details such as hyperparameters. To promote reproducibility and transparency, we detail the key hyperparameters, architectural choices, and implementation decisions used in our benchmark experiments.

\textbf{Partial Observability:}  
Since the pump scheduling problem is naturally partially observable, we employ a Long Short-Term Memory (LSTM) network~\citep{hausknecht15} to enrich the observation space with temporal information. Specifically, we stack four sequential observations using an LSTM layer.

\textbf{Prioritized Experience Replay (PER):}  
Experience Replay~\citep{lin92experience} decorrelates sequential observations by storing transitions and sampling them in mini-batches~\citep{mnih13, mnih2015human}. Rather than sampling uniformly, we adopt Prioritized Experience Replay (PER)~\citep{schaul2016prioritized}, which prioritizes transitions based on their temporal-difference (TD) error $\delta$.

\paragraph{Implementation Choices.}

Following prior work, we used five Q-networks for REM and three for Maxmin Q-learning after empirical evaluation of these ensemble sizes. We adopt the Huber loss for REM, as suggested by its authors, and the Mean Squared Error (MSE) loss for all other algorithms. For the generative model $G_\omega$ used in BCQ, we employ a linear regression model implemented with Scikit-Learn~\citep{scikit-learn}, selecting its hyperparameters via grid search.

The full list of hyperparameters used is provided in Table~\ref{tab:hyperparameters}.

\begin{table}[!ht]
    \centering
  \begin{tabular}{cc}
    \hline
    \textit{Hyperparameter} & \textit{Value}\\
    \hline
    Mini-batch size & 36\\
    (LSTM, Dense, Dense) nodes & (100, 100, 100) \\
    Update target $\lambda$ & 12000\\
    % Loss function (REM) & Huber loss\\
    % Loss function (BCQ) & Mean Squared Error\\
    Optimizer & Adam\\
    Learning rate & 0.00003 \\
    Discount factor & 0.99\\
    $\alpha$ (PER) & 0.6 \\
    $\beta$ (PER) & 0.4 $\rightarrow$ 1 \\
    \#Q-Networks (REM) & 5 \\
    \#Q-Networks (Maxmin Q-learning) & 3 \\
    BCQ threshold $\tau$ & 0.3 \\
    $L_2$ regularization (dense layers) & 0.000001 \\
    Hardware & GPU V100 \\
    \hline
\end{tabular}
\vspace{0.2cm}
 \caption{Hyperparameters and architectural choices used in benchmark experiments.}
 \label{tab:hyperparameters}
\end{table}

\section{Extending the Simulator for Variable-Speed Pumps}\label{sec:continuous}

Speed control is an efficient method for adjusting the operating point of a pump. In this configuration, the motor frequency is varied using a frequency converter, which alters the characteristic curves of the pump, namely the hydraulic head versus flow rate and efficiency versus flow rate relationships. Changes in flow rate and hydraulic head can be described by affinity laws, as given in Equations~\ref{eq:affinity_laws_Q} and~\ref{eq:affinity_laws_H}, where the flow rate ($Q$) varies proportionally to the pump speed ($n$) and the hydraulic head ($H$) varies proportionally to the square of the pump speed:

\begin{equation}\label{eq:affinity_laws_Q}
    Q \propto n
\end{equation}
\begin{equation}\label{eq:affinity_laws_H}
    H \propto n^2
\end{equation}

From Equations~\ref{eq:affinity_laws_Q} and~\ref{eq:affinity_laws_H}, it follows that the hydraulic head changes with the square of the flow rate, as expressed in Equation~\ref{eq:affinity_parabola}, where $Q_1$ and $H_1$ correspond to the flow rate and hydraulic head at the initial reference speed:

\begin{equation}\label{eq:affinity_parabola}
    H_x = \left( \frac{H_1}{Q_1^2} \right) Q_x^2
\end{equation}

\begin{figure}[!ht]
    \centering
    \subfigure[]{\includegraphics[width=.495\textwidth]{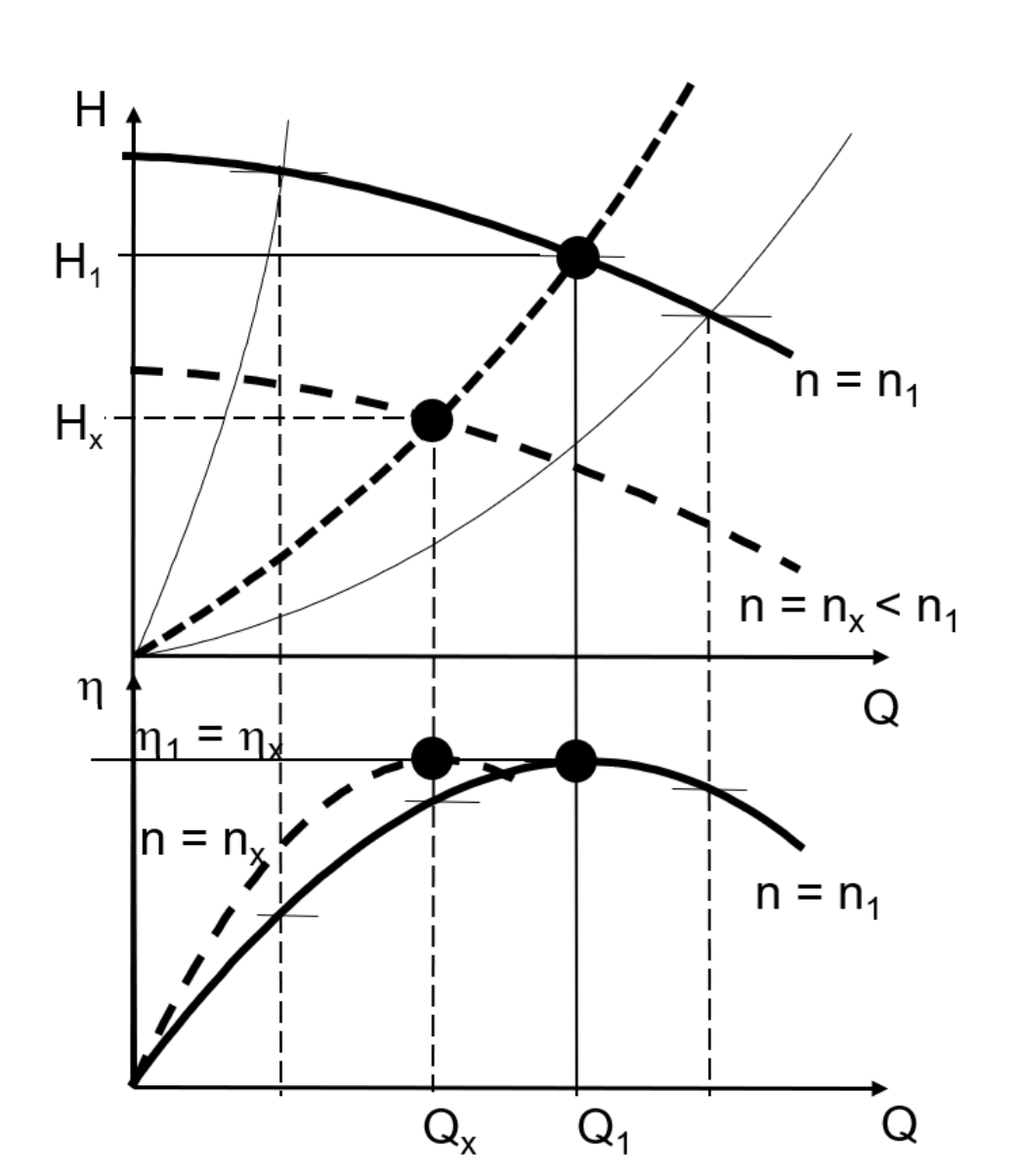}\label{fig:Affinitaet}}
    \subfigure[]{\includegraphics[width=.495\textwidth]{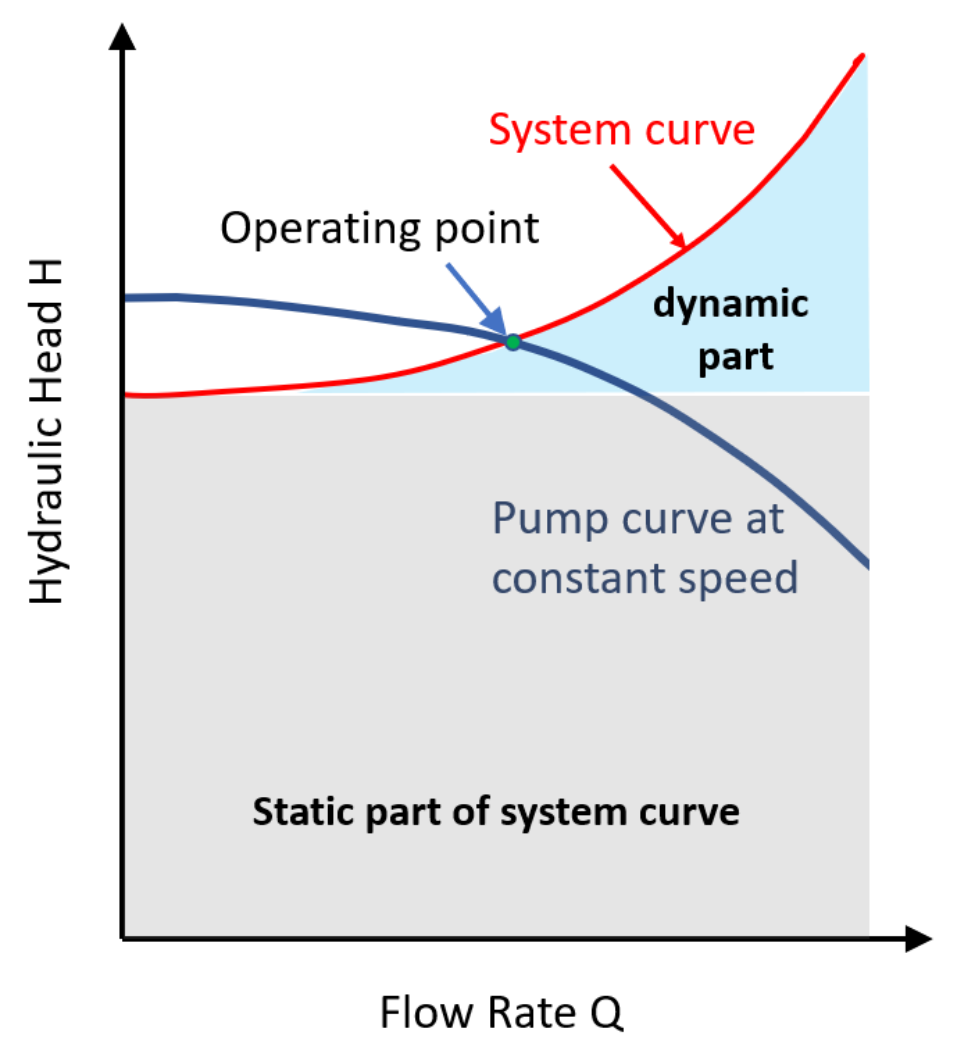}\label{fig:SystemCurve}}
    \caption{(\subref{fig:Affinitaet}) Influence of speed control on the pump curve~\citep{hellmann}. (\subref{fig:SystemCurve}) Relationship between the pump curve and the system curve.}
\end{figure}

Thus, under speed variation, all points on the original pump curve move along parabolic trajectories through the origin (Figure~\ref{fig:Affinitaet}).

Changing the speed also modifies the efficiency curve of the pump. If the operating point corresponds to the point of best efficiency at the nominal speed, it remains the point of best efficiency at any other speed. Other points on the efficiency curve shift along quadratic curves while maintaining their position relative to the best-efficiency point. The maximum efficiency value remains approximately constant. Effects due to variations in the Reynolds number can be incorporated using empirical relationships, such as Ackeret's formula~\citep{pfleiderer}:

\begin{equation}\label{eq:ackeret}
    \frac{\left(1-\eta_{1,\text{opt}}\right)}{\left(1-\eta_{2,\text{opt}}\right)} = \left(1 - V\right) + V \left( \frac{Re_2}{Re_1} \right)^{\left( \frac{1}{\alpha} \right)}
\end{equation}

\begin{figure}[!ht]
    \centering
    \includegraphics[width=1\textwidth]{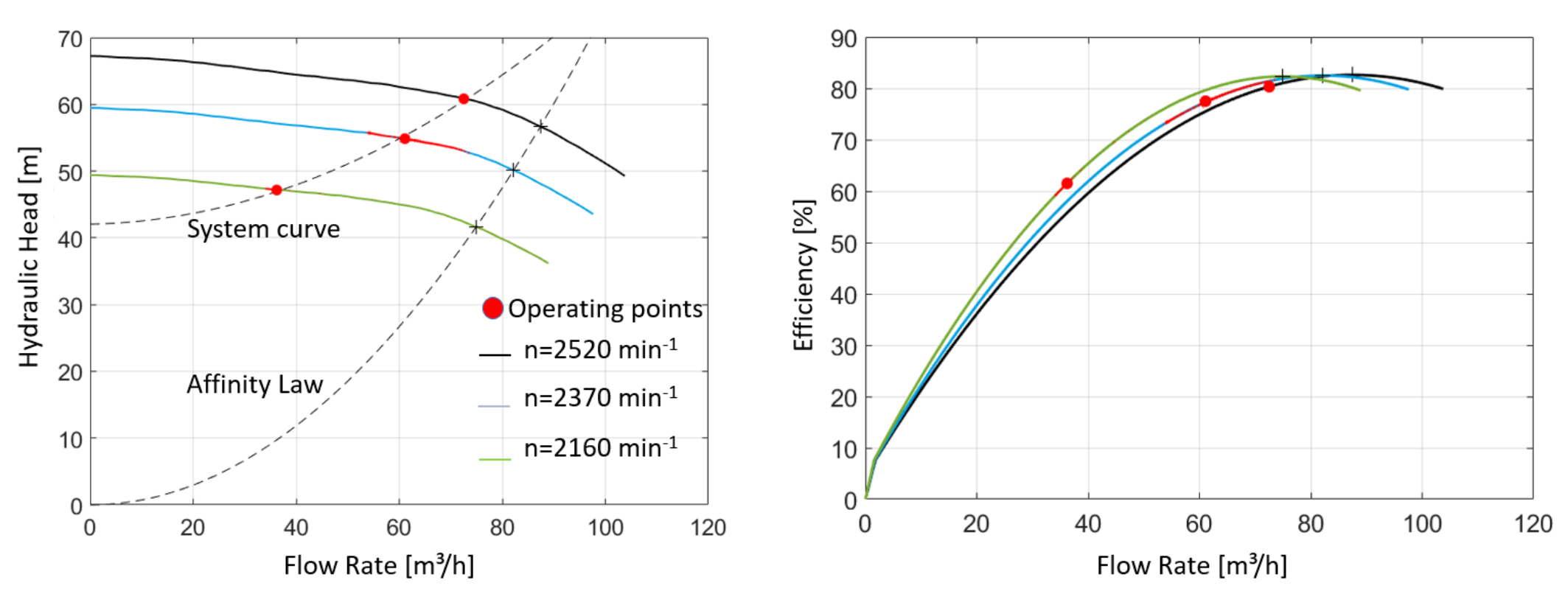} 
    \caption{Effect of speed control on the operating point of a pump with nonzero static head.}
    \label{fig:SpeedControl}
\end{figure}

The change in the pump’s operating point due to speed control depends critically on the shape of the system curve. If the system has no static head (\emph{i.e.}, purely dynamic losses), the system curve coincides with the affinity parabola, and the pump can theoretically operate at its point of best efficiency across a range of speeds. However, in practical systems with a nonzero static head, as illustrated in Figure~\ref{fig:SpeedControl}, the intersection of the pump curve and the system curve deviates from the ideal affinity parabola. Consequently, the pump operates at reduced efficiency under part-load conditions when the speed is lowered.

To enable these effects to be modeled, a speed control function can be incorporated into the main simulator code. This would allow recalculating the pump curve dynamically as the operating speed varies.

\end{document}